\documentclass{article}

     \PassOptionsToPackage{numbers, compress}{natbib}
     \usepackage[preprint]{neurips_2019}

\usepackage[utf8]{inputenc} % allow utf-8 input
\usepackage[T1]{fontenc}    % use 8-bit T1 fonts
\usepackage{hyperref}       % hyperlinks
\usepackage{url}            % simple URL typesetting
\usepackage{booktabs}       % professional-quality tables
\usepackage{amsfonts}       % blackboard math symbols
\usepackage{nicefrac}       % compact symbols for 1/2, etc.
\usepackage{microtype}      % microtypography

\usepackage{amsmath}
\usepackage{amsfonts}
\usepackage{float}
\usepackage{tikz}
\usepackage{xcolor}
\usepackage{colortbl}
\usepackage{adjustbox}
\usepackage{breqn}	
\usepackage{soul}
\usepackage{framed}
\usepackage{multirow}
\usepackage{enumitem}
\usepackage{setspace}
\usepackage{hyperref}
\usepackage{makecell}
\usepackage{calc}
\usetikzlibrary{arrows}
\usetikzlibrary{shapes.geometric}
\usetikzlibrary{shapes.misc, positioning}
\usetikzlibrary{shapes.multipart, shapes.arrows}

\usepackage{xspace}

\newcommand{\ie}{i.e.\@\xspace}

\newcommand{\cf}{cf.\@\xspace}
\newcommand{\etal}{et al.\@\xspace}

\tikzset{
	treenode/.style = {align=center, inner sep=0pt, text centered,
		font=\sffamily},
	tochild/.style={draw,-latex},
	toparent/.style={draw,latex-},
	noedge/.style={draw,latex-, white},
	blacknode/.style = {treenode, regular polygon,regular polygon sides=6, white, draw=black, fill=black, text width=1.5em}, 
	greynode/.style = {treenode, regular polygon,regular polygon sides=6, white, draw=black, fill=black!50, text width=1.5em}, 
	whitenode/.style = {treenode, regular polygon,regular polygon sides=6, black, draw=black, text width=1.5em, very thick}, 
	whitenodefill/.style = {treenode, circle, black, draw=black, fill=white, text width=1.5em}, 
	rednode/.style = {treenode, regular polygon,regular polygon sides=6, red, draw=red, text width=1.5em, very thick}, 
	fullrednode/.style = {treenode, regular polygon,regular polygon sides=6, white, draw=white, fill=red, text width=1.5em, very thick},
	block/.style= {draw, rectangle, minimum width=3cm,minimum height=1cm},
	smallblock/.style= {draw, rectangle, minimum width=2cm,minimum height=0.75cm},
}

\title{Adding Intuitive Physics to Neural-Symbolic Capsules Using Interaction Networks}

\author{%
  Michael Kissner \\
  Institute for Applied Computer Science\\
  Bundeswehr University Munich\\
  \texttt{michael.kissner@unibw.de} \\
  \And
  Helmut Mayer \\
  Institute for Applied Computer Science\\
  Bundeswehr University Munich\\
  \texttt{helmut.mayer@unibw.de} \\
}

\begin{document}

\maketitle

\begin{abstract}
	Many current methods to learn intuitive physics are based on interaction networks and similar approaches. However, they rely on information that has proven difficult to estimate directly from image data in the past. We aim to narrow this gap by inferring all the semantic information needed from raw pixel data in the form of a scene-graph. Our approach is based on neural-symbolic capsules, which identify which objects in the scene are static, dynamic, elastic or rigid, possible joints between them, as well as their collision information. By integrating all this with interaction networks, we demonstrate how our method is able to learn intuitive physics directly from image sequences and apply its knowledge to new scenes and objects, resulting in an inverse-simulation pipeline.
\end{abstract}

\section{Introduction}

An important feature needed for future planning agents is the ability to perform simulations. While it is possible to design this simulation engine by hand, it is far more interesting for the agent to learn it from scratch. This entails learning the intuitive physics of a system and interaction networks \cite{Battaglia:2016} aim to solve this problem. However, the latter requires quite a bit of scene understanding and there is a gap between the information needed and what can be extracted from pixel data using current methods. Especially complex object properties, such as plasticity, angular momentum and joints, are difficult to learn. We, thus, propose a novel method to narrow this gap, by using neural-symbolic capsules \cite{Kissner:2019} to perform inverse-simulation.

We begin by giving some background on neural-symbolic capsules (Section \ref{sec31}), which are a generalization of the original capsule method introduced in \cite{Hinton:2011}\cite{Sabour:2017}. This capsule network follows the "vision as inverse-graphics" principle and produces a scene-graph of the current frame. We propose an extension to the capsule network, so that it works with image sequences and we extract interesting properties from the resulting scene-graphs (Section \ref{sec:Video}), which are vital to perform intuitive physics. All this information is integrated with an interaction network (Section \ref{sec4}). We discuss implementation details and present results of our approach doing physical predictions (Section \ref{sec5}). Finally a conclusion is given and an outlook.

\section{Related Work}

While there are end-to-end proposals for intuitive physics \cite{Lerer:2016}\cite{Ehrhardt:2017}\cite{Iten:2018}, we will focus on interaction networks \cite{Battaglia:2016}, which rely on semantic knowledge of the scene to do physical predictions. Using semantics as part of the process of learning or predicting intuitive physics has proven to be successful and is used in many related methods \cite{Chang:2017}\cite{Wu:2017}\cite{Steenkiste:2018}\cite{Kipf:2018}. Such models for physical scene understanding often follow the idea of an inverse game engine \cite{Battaglia:2013}\cite{Ullman:2017}.

Since the introduction of interaction networks, there has been a lot of research into extending and refining these methods, as well as integrating them with other models. One such extension has been to automatically discover object relations in a scene \cite{Raposo:2017}. An alternative approach has been to connect interaction networks directly with multilayer perceptrons. Watters \etal \cite{Watters:2017} showed that by adding a front-end in the form of a convolutional neural network, intuitive physics can be learned for rigid bodies directly from video data. There has also been research into integrating unsupervised methods \cite{Zheng:2018}, integrating graph networks \cite{Sanchez-Gonzalez:2018}, optimizing the flow of information to and from interaction networks \cite{Hamrick:2017} and using interaction networks in planning systems \cite{Pascanu:2017}.

\section{Neural-Symbolic Capsules}

\subsection{Background}\label{sec31}

We start by giving a short summary of the neural-symbolic capsules introduced in \cite{Kissner:2019}, which are based on \cite{Hinton:2011}\cite{Sabour:2017}. The overall idea of capsules is to increase the amount of information that can be passed between nodes of the network. Where a neuron typically has only a single scalar as its output, a neural-symbolic capsule $\Omega$ produces a vector $\vec{\alpha}_\Omega$. We refer to this vector as the attributes of $\Omega$, such as position and size. As input, the capsule takes the set of attributes $\vec{\alpha}_{1,\cdots,\vert\lambda\vert}$ of its connected capsules $\lambda_i$ and refer to $\Omega$ as the parent of $\lambda_i$. The output is calculated by using a non-linear function $\gamma$, such that $\vec{\alpha}_{\Omega}=\gamma(\vec{\alpha}_{1},\cdots,\vec{\alpha}_{\vert\lambda\vert})$. Neural-symbolic capsules are bi-directional and each capsule also has an inverse function $g$ with $\vec{\alpha}_{1,\cdots,\vert\lambda\vert} = g(\vec{\alpha}_{\Omega})$, where $\gamma$ acts as the encoder and $g$ as the decoder. However, there is no guarantee that $g$ or $\gamma$ are invertible. Instead, given one, we approximate the other by minimizing
\begin{equation}\label{eq:Mini}
||g(\gamma(\vec{\alpha})) - \vec{\alpha}|| \;\;\; .
\end{equation}

Throughout this work we will use $\vec{\alpha}_\Omega^{k_1,\cdots,k_n}$ to describe the subset of components $\alpha_\Omega^{k_1}, \cdots, \alpha_\Omega^{k_n}$ of the attribute vector $\vec{\alpha}_\Omega$. We may also write $\vec{\alpha}_\Omega^{k_1,\cdots,k_n} = \alpha_\Omega^{k_1} \oplus \cdots  \oplus \alpha_\Omega^{k_n}$, where $\oplus$ indicates concatenation. 

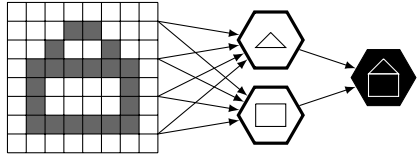
\begin{figure}
	\centering
	\begin{adjustbox}{max width=0.5\textwidth}
		\begin{tikzpicture}

		%- Grammar Parse-Tree
		\node [blacknode] at (0, 0)  {}  
		child[grow=left]
		{ 
			node (A1) [whitenode] at (0, 0.5) {} edge from parent[toparent]
			{}
		}
		child[grow=left]
		{ 
			node (A2) [whitenode] at (0, -0.5) {} edge from parent[toparent]
			{}
		};
		
		\fill[black!60] (-3.25,-0.75) rectangle (-3.5,-0.5);
		\fill[black!60] (-3.25,-0.5) rectangle (-3.5,-0.25);
		\fill[black!60] (-3.25,-0.25) rectangle (-3.5,-0.0);
		\fill[black!60] (-3.25,-0.0) rectangle (-3.5,0.25);
		
		\fill[black!60] (-4.5,-0.75) rectangle (-4.75,-0.5);
		\fill[black!60] (-4.5,-0.5) rectangle (-4.75,-0.25);
		\fill[black!60] (-4.5,-0.25) rectangle (-4.75,-0.0);
		\fill[black!60] (-4.5,-0.0) rectangle (-4.75,0.25);
		
		\fill[black!60] (-3.5,-0.0) rectangle (-3.75,0.25);
		\fill[black!60] (-3.75,-0.0) rectangle (-4,0.25);
		\fill[black!60] (-4.0,-0.0) rectangle (-4.25,0.25);
		\fill[black!60] (-4.25,-0.0) rectangle (-4.5,0.25);
		
		\fill[black!60] (-3.5,-0.75) rectangle (-3.75,-0.5);
		\fill[black!60] (-3.75,-0.75) rectangle (-4,-0.5);
		\fill[black!60] (-4.0,-0.75) rectangle (-4.25,-0.5);
		\fill[black!60] (-4.25,-0.75) rectangle (-4.5,-0.5);
		
		\fill[black!60] (-4.25,0.25) rectangle (-4.5,0.5);
		\fill[black!60] (-3.5,0.25) rectangle (-3.75,0.5);
		\fill[black!60] (-3.75,0.5) rectangle (-4,0.75);
		\fill[black!60] (-4.0,0.5) rectangle (-4.25,0.75);
		
		\draw[scale=0.25] (-12, -4) grid (-20, 4);

		\draw[toparent, black] (A1) -- (-3, 0.75);
		\draw[toparent, black] (A1) -- (-3, 0.25);
		\draw[toparent, black] (A1) -- (-3, -0.25);
		\draw[toparent, black] (A1) -- (-3, -0.75);
		\draw[toparent, black] (A2) -- (-3, 0.75);
		\draw[toparent, black] (A2) -- (-3, 0.25);
		\draw[toparent, black] (A2) -- (-3, -0.25);
		\draw[toparent, black] (A2) -- (-3, -0.75);
		
		\draw [white] (0,0.25) -- (0.2,0.05) -- (-0.2,0.05) -- (0,0.25) ;
		\draw [white] (0.2, 0.05) -- (0.2,-0.25) -- (-0.2, -0.25) -- (-0.2, 0.05) -- (0.2, 0.05);
		
		\draw [black] (-1.5,0.6) -- (-1.7,0.4) -- (-1.3,0.4) -- (-1.5,0.6) ;
		\draw [black] (-1.7, -0.35) -- (-1.7,-0.65) -- (-1.3, -0.65) -- (-1.3, -0.35) -- (-1.7, -0.35);
		
		\end{tikzpicture}
	\end{adjustbox}
	\caption{A capsule network detecting $\left[\textit{wall}\,\right]$, $\left[\textit{roof}\,\right]$ as parts of $\left[\textit{house}\right]$ from an image. Capsules are depicted as hexagons to avoid confusing with neurons.} \label{fig:CapsNet}
\end{figure}

Each capsule has an associated symbol that represents a generalized object which the capsule detects and takes as input only those other capsules that are its parts. For example, the $\left[\textit{house}\right]$ capsule has as its inputs $\left[\textit{wall}\,\right]$ and $\left[\textit{roof}\,\right]$ (\cf Figure \ref{fig:CapsNet}). 

Each capsule representing a part can itself be made up of parts from lower layers. The lowest layer is connected directly to the input pixels and, thus, is made up of capsules that represent graphical primitives, such as a $\left[\textit{square}\right]$ or $\left[\textit{edge}\right]$. We differentiate between these two types and refer to the lowest layer capsules as \textbf{primitive capsules} and all remaining as \textbf{semantic capsules}. 

The primitive capsules are chosen so that the graphical primitives they represent have a known rendering function/decoder $g$. This allows us to perform de-rendering by choosing a regression model for $\gamma$ and train it with synthetic data using Equation \ref{eq:Mini}. Semantic capsules, on the other hand, have a freely definable and known decoder $\gamma$, which we use to train a regression model for $g$.

As each object can be made up of different parts depending on the viewpoint, configuration or style, a routing-by-agreement protocol is devised, based on \cite{Sabour:2017}. For example, a $\left[\textit{house}\right]$ capsule can be made up of $\left[\textit{wall}\,\right]\left[\textit{roof}\,\right]$ or  $\left[\textit{1stfloor}\,\right]\left[\textit{2ndfloor}\,\right]\left[\textit{roof}\,\right]$ or other combinations of input capsules. We refer to each such possibility as a route $r$ and the following protocol is devised to find the best fitting route during a feed-forward pass:

\begin{enumerate}
	
	\item The output $\vec{\alpha}_{\Omega_r}$ for a route $r$ is calculated using
	\begin{equation}\label{eq:CapsuleInternals0}
	\vec{\alpha}_{\Omega_r}=\gamma_r(\vec{\alpha}_{1,\cdots,\vert\lambda\vert})\;\;\; .
	\end{equation}
	
	\item For each input $\vec{\alpha}_{j_r}$ for a route $r$, we estimate the expected input value $\vec{\tilde{\alpha}}_{j_r}$ as if $\vec{\alpha}_{j_r}$ were unknown, using the following equation:
	\begin{equation}\label{eq:CapsuleInternals1}
	\vec{\tilde{\alpha}}_{j} = g_{r,j}(\vec{\alpha}_{\Omega_r})\;\;\; .
	\end{equation}
	
	\item The activation probability $p_{\Omega_r}$ of a route $r$ is calculated as 
	\begin{equation}\label{eq:CapsuleInternals2}
	p_{\Omega_r} = \frac{1}{\vert(\lambda)_r\vert}\;\;\sum_{(\lambda)_r}\; \frac{\lVert Z\left(\vec{\alpha}_i, \vec{\tilde{\alpha}}_i\right)\rVert_1}{\vert Z \vert} \cdot w\left(\frac{p_{i}}{\bar{p}_i} - 1 \right)\;\;\; ,
	\end{equation}
	where $(\lambda)_r$ denotes the set of all inputs that contribute to a route $r$, $p_{i}$ the route's input capsule's probability of activation, $Z$ an agreement-function with output vector of size $\vert Z \vert$, $\lVert \cdot \rVert_1$ the $l_1$-norm, $w$ some window function with $w(0) = 1$, $\sup\{w\} = 1$ and $\bar{p}_i$ the past mean probability for that input. 
	
	\item Steps 1. - 3. are repeated for each $r\in R(\Omega)$.
	
	\item Find the route that was most likely used
	\begin{equation}\label{eq:CapsuleInternals3}
	r_{\textit{final}} = \sup_r \{ p_{\Omega_r} \}
	\end{equation}
	and set the final output as
	
	\begin{align}\label{eq:CapsuleInternals4}
	&p_\Omega = p_{\Omega_{r_{\textit{final}}}} \\
	&\vec{\alpha}_\Omega = \vec{\alpha}_{\Omega_{r_{\textit{final}}}} \;\;\; . 
	\end{align}
	
\end{enumerate}

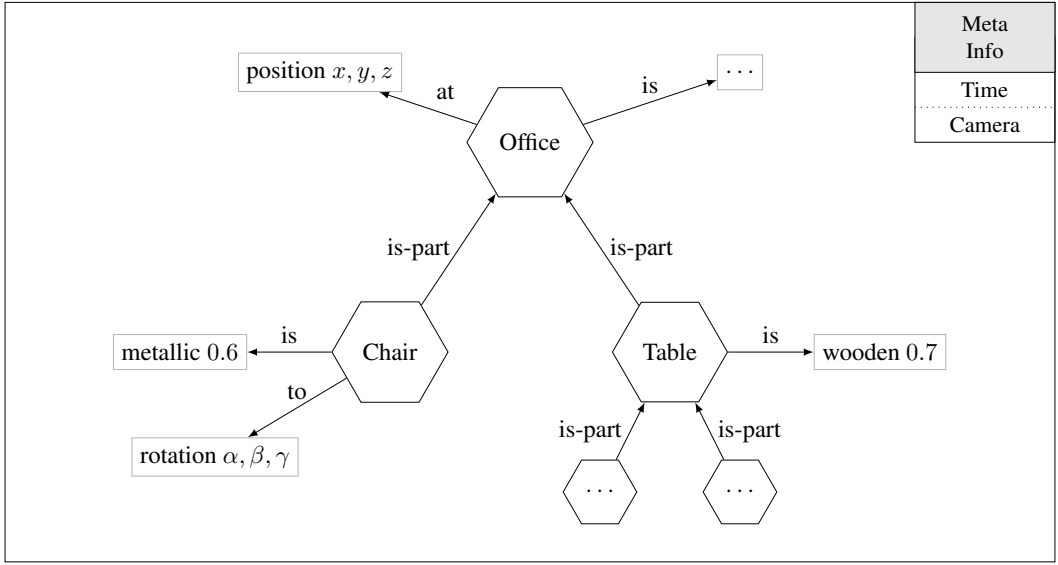
\begin{figure}
	\centering
	\begin{adjustbox}{max width=1\textwidth}
		\begin{tikzpicture}		
		
		%\node (Box) [draw, minimum width=15cm, minimum height=8cm] at (0.0, -1.0) {};
		%\node (MetaBox) [draw, minimum width=2cm, minimum height=2.5cm] at (6.5, 1.75) {};
		%\node (MetaText) [draw, fill=black!10!white, minimum width=2cm, minimum height=1cm, text width=1.7cm, align=center] at (6.5, 2.5) %{\footnotesize{Meta\\Info}};
		%\node (MetaText) [minimum width=2cm, minimum height=0.5cm, text width=1.7cm, align=center] at (6.5, 1.75) {\footnotesize{Type}};
		%\draw [draw, dotted] (5.5,1.5) -- (7.5,1.5);
		%\node (MetaText) [minimum width=2cm, minimum height=0.5cm, text width=1.7cm, align=center] at (6.5, 1.25) {\footnotesize{Time}};
		%\draw [draw, dotted] (5.5,1.0) -- (7.5,1.0);
		%\node (MetaText) [minimum width=2cm, minimum height=0.5cm, text width=1.7cm, align=center] at (6.5, 0.75) {\footnotesize{Camera}};

		\node (Box) [draw, minimum width=15cm, minimum height=8cm] at (0.0, -1.0) {};
		\node (MetaBox) [draw, minimum width=2cm, minimum height=1.5cm] at (6.5, 1.75) {};
		\node (MetaText) [draw, fill=black!10!white, minimum width=2cm, minimum height=1cm, text width=1.7cm, align=center] at (6.5, 2.5) {\footnotesize{Meta\\Info}};
		\node (MetaText) [minimum width=2cm, minimum height=0.5cm, text width=1.7cm, align=center] at (6.5, 1.75) {\footnotesize{Time}};
		\draw [draw, dotted] (5.5,1.5) -- (7.5,1.5);
		\node (MetaText) [minimum width=2cm, minimum height=0.5cm, text width=1.7cm, align=center] at (6.5, 1.25) {\footnotesize{Camera}};
		
		\node (Office) [regular polygon,regular polygon sides=6, draw, minimum width=0.5cm, minimum height=0.5cm] at (0.0, 1.0) {Office};		
		\node (Table) [regular polygon,regular polygon sides=6, draw, minimum width=0.5cm, minimum height=0.5cm] at (2.0, -2.0) {Table};
		\node (Chair) [regular polygon,regular polygon sides=6, draw, minimum width=0.5cm, minimum height=0.5cm] at (-2.0, -2.0) {Chair};
		\node (TableP1) [regular polygon,regular polygon sides=6, draw, minimum width=0.5cm, minimum height=0.5cm] at (3.0, -4.0) {$\cdots$};
		\node (TableP2) [regular polygon,regular polygon sides=6, draw, minimum width=0.5cm, minimum height=0.5cm] at (1.0, -4.0) {$\cdots$};
		
		\path [tochild] (Table) -- node [right,midway] {is-part} (Office);
		\path [tochild] (Chair) -- node [left,midway] {is-part} (Office);
		\path [tochild] (TableP1) -- node [right,midway] {is-part} (Table);
		\path [tochild] (TableP2) -- node [left,midway] {is-part} (Table);
		%\path [draw,latex'-latex'] (Chair) -- node [above,midway] {collides} (Table);

		\node (Adj1) [draw=black!30!white, minimum width=0.5cm, minimum height=0.5cm] at (5.0, -2.0) {wooden $0.7$};
		\node (Adj2) [draw=black!30!white, minimum width=0.5cm, minimum height=0.5cm] at (-5.0, -2.0) {metallic $0.6$};
		\node (Adj3) [draw=black!30!white, minimum width=0.5cm, minimum height=0.5cm] at (3.0, 2.0) {$\cdots$};
		\node (Prep1) [draw=black!30!white, minimum width=0.5cm, minimum height=0.5cm] at (-3.0, 2.0) {position $x,y,z$};
		\node (Prep2) [draw=black!30!white, minimum width=0.5cm, minimum height=0.5cm] at (-4.5, -3.5) {rotation $\alpha,\beta,\gamma$};
		
		\path [tochild] (Chair) -- node [above,midway] {is} (Adj2);
		\path [tochild] (Table) -- node [above,midway] {is} (Adj1);
		\path [tochild] (Office) -- node [above,midway] {is} (Adj3);
		\path [tochild] (Office) -- node [above right,midway] {at} (Prep1);
		\path [tochild] (Chair) -- node [above,midway] {to} (Prep2);
		
		\end{tikzpicture}
	\end{adjustbox}
	\caption{Example scene-graph for an office scene extracted from capsule observation tables, collision data and meta-information.} \label{fig:SemanticNetwork}
\end{figure}

This routing protocol represents the internals of each capsule, allowing us to connect them to form a capsule network. Even though this network has a top-most layer, this is not the output layer. Instead, each individual capsule itself is an output. After the feed-forward pass, the activation of each individual capsule and its relative position in the network allows us to form a scene-graph as in Figure \ref{fig:SemanticNetwork}.

The capsule network can also be used feed-backward. In this direction, the network acts as a \textit{parse-tree} of an attributed generative grammar, where we translate capsules $\to$ symbols and routes $\to$ rules. A starting point in the tree is chosen and by successively applying the decoders $g$ of the individual capsules, a rendered image is produced.

\subsection{Moving Towards Video}\label{sec:Video}

For static scenes and objects, the attributes of individual capsules are either interpreted as an \textbf{adjective} (metallic, wooden) or a \textbf{preposition} (position, rotation, size). Once we move towards video input and dynamic scenes, a third interpretation arises that describes movements and poses, the \textbf{verb} attribute (walking, folding).

To detect movement and change of pose, we must track objects across frames. For this we assume that the camera moves smoothly. Perfect tracking would entail being able to predict positions, which in turn requires knowledge of the physics in the scene. We run into a deadlock, as we have no way of inferring physics, without prior tracking. Thus, we rely on a rudimentary tracking algorithm based on velocities and similarities of the attributes across frames, ignoring higher order effects. We can, however, always include force contributions once they are known.

By $\Lambda_i$ we denote the set of all observations of some symbol/capsule $\lambda_i$ the capsule network has made in an image. $\{\Lambda_{1,t}, \Lambda_{2,t}, \cdots \}$ then denotes the set of all objects in a frame at time $t$, split into their respective symbol types. By $(\overrightarrow{{\Lambda}_{i,t}})_a$ we denote the observed attributes of the $a$th object in $\Lambda_i$. Our goal is to get an estimate for the relations between the objects in sets $\Lambda_{i,t-1}$ and $\Lambda_{i,t}$, \ie, object continuity. This allows us to calculate important quantities, such as the velocity of the attributes, $\frac{\delta (\overrightarrow{{\Lambda}_{i,t}})_a}{\delta t}$. 

Let $(a, b)$ denote a relation of the $a$th object at time $t-1$ with the $b$th object at time $t$ and we allow the possibility of $(a, \cdot)$ or $(\cdot,b)$ not having a partner. We find the best set of similarity relations $P_{i,\textit{sim}}$ out of all possible relations $\mathcal{P}_i$ between the observations $\Lambda_{i,t-1}$ and $\Lambda_{i,t}$ by minimizing:
\begin{equation}\label{eq:continue}
\min_{P_{i,j}\in \mathcal{P}_i, \textbf{R}_t, \vec{x}_t} \left\{ \sum_{(a,b) \in P_{i,j}} \left\lVert \vec{w} \cdot \left( (\overrightarrow{{\Lambda}_{i,t-1}})_a + \frac{\delta (\overrightarrow{{\Lambda}_{i,t-1}})_a}{\delta t}\Delta t - \left[\textbf{R}_t\cdot(\overrightarrow{{\Lambda}_{i,t}})_b + \vec{x}_t \right] \right) \right\lVert \right\} \;\;\;,
\end{equation}
where $\textbf{R}_t$ denotes the camera rotation matrix that only acts on the position and rotation attributes and $\vec{x}_t$ the camera translation that only acts on the position attributes. The final $P_{i,\textit{sim}}$ that minimizes Equation \ref{eq:continue} gives us a map between the objects of the two frames. $\textbf{R}_t$ and $\vec{x}_t$ describe how the scene and camera have moved from the preceding frame to the new one. While we minimize over all attributes, some (such as adjectives) give a better indication of continuity than others (such as verbs) and introduce the hyperparameter $\vec{w}$ in Equation \ref{eq:continue} to allow for fine-grained regularization. 

The error grows in the order of $(\Delta t)^2$ and, thus, we can reduce it by choosing smaller time-steps. Reducing $\Delta t$ also minimizes the effects of the camera and velocity in Equation \ref{eq:continue} as $\textbf{R}_t \to \textbf{I}$ and $\vec{x}_t \to \vec{0}$, leaving the pure similarity search of complexity $\mathcal{O}(n!)$ as the dominant factor. 

To find a solution to Equation \ref{eq:continue}, consider a scene with a dining table, four chairs and 50 balls on the table. In two frames with different viewpoints, it is easy to spot the relation frame-to-frame for the table, but difficult for the balls. Our approach is to first find relations for objects that are few in the scene and move up to objects that are plentiful, updating our knowledge of our own orientation each step.

We begin with the smallest set with size $\vert \Lambda_{k,t} \vert = \inf_i \{  \vert \Lambda_{i,t}  \vert \}$ and perform a full search to find $P_{k,\textit{sim}}$ and an initial estimate for $\textbf{R}_t$ and $\vec{x}_t$. Next, we focus on larger sets $\Lambda_{i,t}$ and reuse the camera attributes as a prior to speed up our search, by making an initial guess for the state each object should be in: 
\begin{equation}\label{eq:nextStep1}
\tilde{(\overrightarrow{{\Lambda}_{i,t}})}_b = \textbf{R}^{-1}_t \cdot  \left[(\overrightarrow{{\Lambda}_{i,t-1}})_a + \frac{\delta (\overrightarrow{{\Lambda}_{i,t-1}})_a}{\delta t}\Delta t - \vec{x}_t \right] \;\;\;.
\end{equation}
We then find relations using 
\begin{equation}\label{eq:nextStep2}
\lVert (\overrightarrow{{\Lambda}_{i,t}})_b - (\tilde{\overrightarrow{{\Lambda}_{i,t}}})_b \lVert < \epsilon(\Delta t) \;\;\;,
\end{equation}
where $\epsilon(\Delta t)$ is a measure for error and the maximal acceleration we expect to happen between frames, before we consider it unphysical. Using the resulting set of relations, we refine our camera attributes and repeat the process until we have found all $P_{i,\textit{sim}}$ for all object types. This effectively reduces the complexity to $\mathcal{O}(n^2)$.

All outliers, \ie objects that have no relation according to Equation \ref{eq:nextStep2}, describe objects that have either newly entered or left the scene, such as through occlusion. To regain continuity for all these outliers, we attempt to reestablish their relation by matching them to lost objects across larger time-steps. We are able do this, as we have access to the camera transformations of intermediary time-steps. Now, should an object re-enter the scene after leaving it or being occluded, tracking can be resumed, if it was not affected by large forces during occlusion. We may, however, always refine our tracking and incorporate acceleration in Equation \ref{eq:continue} and \ref{eq:nextStep1} using the results from intuitive physics given below.

Another important aspect of tracking individual objects is to train verb attributes. Each frame may describe a pose in a sequence of poses. This pose animation is key-framed by a verb attribute in the range of $[0,1]$. A previously unseen pose is trained according to the procedure outlined in \cite{Kissner:2019} and follows that of adjective attributes. The main difference for training verbs is, that we transform the previously trained range of the attribute to $[0, 1-\epsilon]$, before training the new pose with value $1$. This is akin to appending a new frame. 

\section{Interactable Neural-Symbolic Capsules}\label{sec4}

\subsection{Background}

We give a short summary of the ideas behind interaction networks \cite{Battaglia:2016} and our notation. The goal of interaction networks is a learnable model for intuitive physics. First, a set of objects $O$ which can interact physically is defined. Generally, not every object $o\in O$ interacts with every other object at the same time. For objects that do interact at a given point in time, the interaction relation is given by the triplet $\langle o_i, o_j, \rho \rangle$, where $o_i$ is the \textit{sender} object, $o_j$ the \textit{receiver} object and $\rho$ describes the details of the relation. If the interaction is bi-directional, a second relation is defined, but with reversed sender and receiver roles. 

Next, an intermediate value $e$ is introduced. It describes the effect of a single interaction, \ie, a single relation triplet, on the overall future of an object. This is especially important when an object is the receiver of multiple interactions. We find an effect $e_k$ for each triplet $\langle o_i, o_j, \rho \rangle_k$ using a regression model $\phi_R$, the \textit{relational model}:
\begin{equation}
\{e_k\} = \phi_R \left[\{\langle o_i, o_j, \rho \rangle_k\}\right] \;\;\; .
\end{equation}

To predict how an object $o_j$ evolves, all $e_k$ that influence it must be aggregated. However, apart from interactions with each other, external effects, such as gravity, might also have an influence. We describe these external effects in the set $X=\{x_j\}$. Next, an aggregation function $a(\cdot)$ is defined, that merges all effects for an object and combines them with the external forces. It serves as input for the final regression model $\phi_O$, the \textit{object model}, that predicts the next time-step:
\begin{equation}
\{o_j^{t+1}\} = \phi_O \left[a(\{\langle o_i^t, o_j^t, \rho \rangle_k\}, \{x_j\}, \{e_k\} )\right] \;\;\; .
\end{equation}

Now, to make predictions, we only need to train $\phi_R$ and $\phi_O$.

\subsection{Collision Detection}

An important measure that we require before we continue with interaction networks is the distance and collision between objects, which we find using the following algorithm: 

\begin{enumerate}
	\item Render objects $A$ and $B$ using the parse-tree.
	\item Find the minimal distance between the pixels (2D) or voxels (3D) of the two renderings using
	\begin{equation}\label{eq:continue3}
	d_{A,B} = \inf\left\{ d(\vec{a}_i^{pos} - \vec{b}_j^{pos}) \;\lvert \; \vec{a}_i \in A,\; \vec{b}_j \in B; \;  a_i^{int} > 0,\; b_j^{int} > 0 \right\} \;\;\; ,
	\end{equation}
	where $\vec{a}, \vec{b}$ are the attributes of individual pixels or voxels, $pos$ and $int$ their positions and intensities and $A,B$ the set of all pixels or voxels for each objects. 
	\item Calculate the boundary normals $\vec{N}_A$ and $\vec{N}_B$ for the points of minimal distance between $A$ and $B$.
\end{enumerate}

We are deliberately vague in regard to calculating the normals, as we have many options. We can infer the normals approximately using the positions of nearby pixels/voxels or we can calculate the normals analytically, as the primitive capsules have a known rendering function $g$.

\subsection{Finding Relations}

To derive relations of the form $\langle o_i, o_j, \rho \rangle$, we must find objects in our scene-graph that are interactable. An important characteristic of our graph is the presence of verb attributes. A verb allows an object's parts to move either independently of each other or constrained by a joint. To make the relation between verbs and the movements they cause explicit, we begin by analyzing the known (\ie, observed in the past) object behavior in our capsule network. We assume $g$ is continuous for all semantic capsules, which is true, for example, for neural networks with a continuous activation function. Our search for interactable objects and their properties proceeds as follows:

\begin{enumerate}
	\item Find a semantic capsule $\Omega$ that has verb attributes $\vec{\alpha}_\Omega^{k_1,\cdots,k_n}$ and one or more child capsules $\lambda_i$ that do not have verb attributes. All child capsules $\lambda_i$ that satisfy this condition form the set of interactable objects $O$.
		
	\item Vary $\vec{\alpha}_\Omega^{k_1,\cdots,k_n}$ in the range $[0,1]^n$ using $g(\vec{\alpha}_\Omega) = \vec{\alpha}_{1,\cdots,\vert\lambda\vert}$ to find the spaces for scale, pose (position and rotation) and collision
	\begin{equation}
	{\mathbf{S}}_i = \{ \vec{\alpha}_i^{\mathit{size}}\} \;\;\; ,
	\end{equation}
	\begin{equation}
	{\mathbf{Q}}_i = \{ \vec{\alpha}_i^{\mathit{pos}} \oplus \vec{\alpha}_i^{\mathit{rot}} \} \;\;\; ,
	\end{equation}
	\begin{equation}
	{\mathbf{Q}}_{ij} = \{ (\textbf{R}_i \cdot (\vec{\alpha}_j^{\mathit{pos}} - \vec{\alpha}_i^{\mathit{pos}})) \oplus (\vec{\alpha}_j^{\mathit{rot}} - \vec{\alpha}_i^{\mathit{rot}}) \}  \;\;\; ,
	\end{equation}
	\begin{equation}
	\mathbf{D}_{ij} = \{ d(\lambda_i, \lambda_j) \}  \;\;\; ,
	\end{equation}
	where $\textbf{R}_i$ is the Euler rotation matrix for $-\vec{\alpha}_i^{\mathit{rot}}$, which is used to move the two objects into the same reference frame. ${\mathbf{Q}}_{ij}$ is also defined for two capsules, even if they do not share a common parent, by finding the next common ancestor instead and successively applying $g$.
	
	\item For the space ${\mathbf{S}}_i$ we define a pseudo-dimension $D_{S_i}$. Consider a space $\hat{\mathbf{S}}_i$, for which there exists a continuous, surjective map $\psi \colon \mathbf{S}_i \to \hat{\mathbf{S}}_i$, such that for every point $p\in{\mathbf{S}}_i$ we have 
	\begin{equation}\label{eq:contract}
	\Vert p - \psi(p) \Vert < \epsilon \;\;\; ,
	\end{equation}
	where $\Vert \cdot \Vert$ measures the Euclidean distance. We are essentially contracting ${\mathbf{S}}_i$ into $\hat{\mathbf{S}}_i$. Note that some neighborhoods $\hat{\mathbf{S}}_{i,j} \subseteq \hat{\mathbf{S}}_i$ are homeomorphic to $\mathbb{R}^{n_j}$. We now let the pseudo-dimension $D_{S_i}$ be $D_{S_i} = \min_\psi [ \max_j n_j ]$, where we attempt to find the best contraction $\psi$ with minimal "dimension" $\max_j n_j$ possible. We do this to remove small numerical errors that accidentally generate unnecessary dimensions and repeat this process for ${\mathbf{Q}}_i$ and ${\mathbf{Q}}_{ij}$.	
\end{enumerate}

We infer characteristics about the individual capsules using the following conditions:

\begin{enumerate}
	\item[A] A verb-less child $\lambda_i$ is a \textbf{rigid body} if $D_{S_i} = 0$.\\
	\textit{$\to$ The size has never been observed to change over time.}
	\item[B] A verb-less child $\lambda_i$ is an \textbf{elastic/plastic body} if $D_{S_i} > 0$.\\
	\textit{$\to$ The size has been observed to change over time.}
	\item[C] A verb-less child $\lambda_i$ is \textbf{static} if $D_{Q_i} = 0$.\\
	\textit{$\to$ The position or rotation has never been observed to change over time.}
	\item[D] A verb-less child $\lambda_i$ is \textbf{dynamic} if $D_{Q_i} > 0$.\\
	\textit{$\to$ The position or rotation has been observed to change over time.}
	\item[E] Two verb-less descendants $\lambda_i$ and $\lambda_j$ are connected by a \textbf{joint} if they collide ($\sup\mathbf{D}_{ij} < \epsilon$) and their total degrees-of-freedom $D_{Q_{ij}}$ in relation to each other is $3 > D_{Q_{ij}} > 0$ (2D) or $6 > D_{Q_{ij}} > 0$ (3D).\\
	\textit{$\to$ The two objects have been observed to move in relation to each other over time in a constrained way.}
\end{enumerate}

For example, consider a 3-dimensional $\left[\textit{house-door}\,\right]$ with a single verb attribute (open) and two children: $\left[\textit{frame}\right]$ and $\left[\textit{door}\,\right]$. According to (1.) of our algorithm, the generalized object $\left[\textit{house-door}\,\right]$ is not interactable, as it has a verb, whereas its parts are interactable, as they do not have verbs. Every time the door opens and closes, the configuration space $\mathbf{Q}_{\mathit{door},\mathit{frame}}$ traces out a curve. Further, the $\left[\textit{frame}\right]$ and $\left[\textit{door}\,\right]$ are in constant collision at the hinges. These conditions are sufficient to identify them as two rigid bodies connected by a joint ($D_{Q_{ij}} = 1$), where $\left[\textit{frame}\right]$ is static ($D_{Q_i} = 0$) and $\left[\textit{door}\,\right]$ is dynamic ($D_{Q_j} = 1$). We, thus, find the expected one degree-of-freedom of a hinged door. This shouldn't be surprising, as a single verb can't parameterize anything higher-dimensional than a curve in configuration space. However, consider a second verb: slam. By varying in the range $[0,1]^2$, we could assume that it traces out a plane, but instead it still describes the same curve in $\mathbf{Q}_{\mathit{door},\mathit{frame}}$, preserving the expected result of one degree-of-freedom.

We now define the set of sender-receiver-relation triplets that we add for all the objects. Each relation is defined by the five previously identified characteristics. We encode all needed properties as shown in Table \ref{tab:relation} and include all the symbol names, attribute values and attribute names for the objects, as they encode more than just position, rotation, scale and velocity. For example, a $\left[\textit{plate}\right]$ with a non-zero metallic attribute might be affected by magnetism, whereas a non-metallic $\left[\textit{plate}\right]$ is not. Now, with all objects and relations in place, we are ready to train relation model $\phi_R$. 

{
\renewcommand{\arraystretch}{2}
\begin{table}
	\caption{\label{tab:relation} Full relation triplet used as input for $\phi_R$.}
	\begin{center}
		\resizebox{\textwidth}{!}{
		\begin{tabular}{ c  c  c  c  c  c  c  c  c  c  c  c  c  c } 
			\hline
			\multicolumn{5}{c}{\cellcolor{green!10!white} Sender $o_i$} & \multicolumn{5}{c}{\cellcolor{blue!10!white} Receiver $o_j$} & \multicolumn{4}{c}{\cellcolor{red!10!white} Relation $\rho$} \\
			
			\hline
			$\lambda_i$ & $\vec{\alpha}_i$ & $\frac{d \vec{\alpha}_i}{d t}$ & \makecell{static / \\ dynamic} & \makecell{rigid / \\ elastic} &  
			$\lambda_j$ & $\vec{\alpha}_j$ & $\frac{d \vec{\alpha}_j}{d t}$ & \makecell{static / \\ dynamic} & \makecell{rigid / \\ elastic} &
			$d_{o_i, o_j}$ & joint & $\vec{N}_{o_i}$ & $\vec{N}_{o_j} $ \\
			\hline

		\end{tabular}}
	\end{center}
\end{table}
}

To train the object model $\phi_O$, we require aggregating all the results from $\phi_R$. This is done by summing the resulting effect vectors $e_k$ and concatenating these with external effects $X$ and all the receiver information highlighted in Table \ref{tab:aggregate}.

{
\renewcommand{\arraystretch}{2}
\begin{table}
	\caption{\label{tab:aggregate} Aggregated data $a(\cdot)$ used as input for $\phi_O$.}
	\begin{center}
		\begin{tabular}{ c  c  c  c  c  c  c } 
			\hline
			\multicolumn{5}{c}{\cellcolor{blue!10!white} Receiver $o_j$} & \multicolumn{2}{c}{\cellcolor{yellow!10!white} Effects}  \\
			
			\hline
			$\lambda_i$ & $\vec{\alpha}_i$ & $\frac{d \vec{\alpha}_i}{d t}$ & \makecell{static / \\ dynamic} & \makecell{rigid / \\ elastic} &  
			$\Sigma e_k$ & $X$ \\
			\hline
								
		\end{tabular}
	\end{center}
\end{table}
}

\section{Implementation and Results}\label{sec5}

We implement the capsule network feeding the interaction network according to \cite{Kissner:2019} and our code for VividNet is found on Github at \url{https://github.com/Kayzaks/VividNet}. For geometric primitives the distance functions are known and we implement collision detection, as well as normal calculation, analytically. Tracking is performed using the methods described in Section \ref{sec:Video}.

Next, we focus on calculating the dimensions of $\textbf{S}_i, \textbf{Q}_i, \textbf{Q}_{ij}$ and $\textbf{D}_{ij}$. Instead of varying every $\alpha_\Omega^{k_1,\cdots,k_n}$ over the entirety of $[0,1]^n$, we use a Monte-Carlo approach. We randomly select a small $n$-dimensional cube $[x_1, x_1+\Delta x]\times\cdots\times[x_n, x_n+\Delta x]\subseteq [0,1]^n$ and map its corners using $g$ to obtain a transformed volume $\hat{V}$. We contract the volume using Equation \ref{eq:contract} to find $V$ and dimension $D_V$. By repeating this process multiple times we determine an estimate for the true dimension from the distribution of the values for $D_V$. We choose the maximum of the distribution as the actual dimension and assume that any outliers were removed by a suitable choice of $\epsilon$ in Equation \ref{eq:contract}.

For simplicity, our implementation of $\phi_R$ and $\phi_O$ relies on manually adjusted designed neural networks and manually adjusted hyper-parameters. We train $\phi_R$ and $\phi_O$ by presenting the interaction network with multiple synthetic scenes involving the collision of two objects. The symbol names, the dimensions $D_{S_i}$, $D_{Q_i}$, as well as the degrees-of-freedom $D_{Q_{ij}}$ are encoded in binary vectors, with one component for each dimension (\cf Table \ref{tab:relation} and \ref{tab:aggregate}). The attribute names are encoded by keeping the position of each attribute type static in the triplet and aggregate.

\begin{figure}
	\centering
	\begin{adjustbox}{max width=1.0\textwidth}
		\begin{tikzpicture}
		\node[inner sep=0pt] at (3,0) {\includegraphics[width=.70\textwidth]{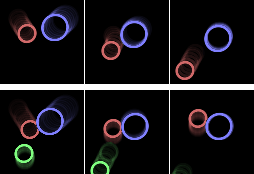}};
		\end{tikzpicture}
	\end{adjustbox}
	\caption{Top shows the prediction of two circles interacting. Bottom shows the prediction of three circles interacting, whereas the interaction network was only trained on two. Movement is illustrated using color and object trails.} \label{fig:results1}
\end{figure}

In our first result, we perform a simple prediction by showing the network a sequence of circles prior to their moment of interaction and asking it to predict future time-steps, where we expect to see an elastic collision. From the two frames, the capsule network is able to infer the speed of both circles and successively predict each frame thereafter, as shown in the top row of Figure \ref{fig:results1}. Our predictions remain plausible up to $\sim 50$ frames, but suffer from accumulated numerical errors thereafter. We then increase the number of circles in the scene and find that the network is able to predict interactions for any number of objects (\cf bottom row of Figure \ref{fig:results1}).

\begin{figure}
	\centering
	\begin{adjustbox}{max width=1.0\textwidth}
		\begin{tikzpicture}
		\node[inner sep=0pt] at (0,0) {\includegraphics[width=.70\textwidth]{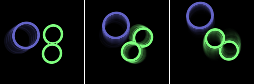}};
		\end{tikzpicture}
	\end{adjustbox}
	\caption{Interaction of a circle with a figure-eight, which is only able to rotate around its axis. Movement is illustrated using color and object trails.} \label{fig:results2}
\end{figure}

For our final experiment, we consider a figure-eight that is unable to move in the x- and y-direction, but able to rotate. It acts as a windmill and we find $\textbf{Q}_i$ only spans one dimension ($D_{Q_i} = 1$), namely rotation. We launch a circle at this figure-eight and predict its response. In Figure \ref{fig:results2} we see, that $\phi_R$ and $\phi_O$ do indeed predict plausible rotations of the figure-eight once it interacts with the circle. 

\section{Conclusion and Outlook}

In this paper we have presented our method for extracting information from pixel or voxel data to perform intuitive physics for simulations. We have shown that interaction networks are able to learn complex physical interactions with different geometries and constraints, using the semantics and features generated by the neural-symbolic capsule network.

By taking the idea of an inverse game-engine literally, we believe that the intuitive physics in our proposed inverse simulation pipeline can be extended to learn more complex symbolic interactions, such as the logic of a game. Feed-forward, it then learns how the game works and, feed-backward, it functions as the actual game engine itself, where the capsule network renders all the graphics with simulated physics and game logic.

\bibliographystyle{unsrt}
\bibliography{egbib}

\begin{thebibliography}{10}

\bibitem{Battaglia:2016}
Peter Battaglia, Razvan Pascanu, Matthew Lai, Danilo~Jimenez Rezende, and Koray
  Kavukcuoglu.
\newblock Interaction networks for learning about objects, relations and
  physics.
\newblock {\em NIPS}, 2016.

\bibitem{Kissner:2019}
Michael Kissner and Helmut Mayer.
\newblock A neural-symbolic architecture for inverse graphics improved by
  lifelong meta-learning.
\newblock {\em arXiv:1905.08910}, 2019.

\bibitem{Hinton:2011}
Geoffrey~E Hinton, Alex Krizhevsky, and Sida~D Wang.
\newblock Transforming auto-encoders.
\newblock {\em International Conference on Artificial Neural Networks}, pages
  44--51, 2011.

\bibitem{Sabour:2017}
Sara Sabour, Nicholas Frosst, and Geoffrey~E Hinton.
\newblock Dynamic routing between capsules.
\newblock {\em NIPS}, 2017.

\bibitem{Lerer:2016}
Adam Lerer, Sam Gross, and Rob Fergus.
\newblock Learning physical intuition of block towers by example.
\newblock {\em ICML}, 2016.

\bibitem{Ehrhardt:2017}
Sebastien Ehrhardt, Aron Monszpart, Niloy~J. Mitra, and Andrea Vedaldi.
\newblock Learning a physical long-term predictor.
\newblock {\em arXiv:1703.00247}, 2017.

\bibitem{Iten:2018}
Raban Iten, Tony Metger, Henrik Wilming, Lidia~del Rio, and Renato Renner.
\newblock Discovering physical concepts with neural networks.
\newblock {\em arXiv:1807.10300}, 2018.

\bibitem{Chang:2017}
Michael~B. Chang, Tomer Ullman, Antonio Torralba, and Joshua~B. Tenenbaum.
\newblock A compositional object-based approach to learning physical dynamics.
\newblock {\em ICLR}, 2017.

\bibitem{Wu:2017}
Jiajun Wu, Joschua~B. Tenenbaum, and Pushmeet Kohli.
\newblock Neural scene de-rendering.
\newblock {\em CVPR}, 2017.

\bibitem{Steenkiste:2018}
Sjoerd~van Steenkiste, Michael Chang, Klaus Greff, and Jürgen Schmidhuber.
\newblock Relational neural expectation maximization: Unsupervised discovery of
  objects and their interactions.
\newblock {\em ICLR}, 2018.

\bibitem{Kipf:2018}
Thomas Kipf, Ethan Fetaya, Kuan-Chieh Wang, Max Welling, and Richard Zemel.
\newblock Neural relational inference for interacting systems.
\newblock {\em ICML}, 2018.

\bibitem{Battaglia:2013}
Peter~W. Battaglia, Jessica~B. Hamrick, and Joshua~B. Tenenbaum.
\newblock Simulation as an engine of physical scene understanding.
\newblock {\em Proceedings of the National Academy of Sciences},
  110(45):18327–18332, 2013.

\bibitem{Ullman:2017}
Tomer~D. Ullman, Elizabeth Spelke, Peter Battaglia, and Joshua~B. Tenenbaum.
\newblock Mind games: Game engines as an architecture for intuitive physics.
\newblock {\em Trends in Cognitive Science}, 21(9):649--665, 2017.

\bibitem{Raposo:2017}
David Raposo, Adam Santoro, David Barrett, Razvan Pascanu, Timothy Lillicrap,
  and Peter Battaglia.
\newblock Discovering objects and their relations from entangled scene
  representations.
\newblock {\em ICLR}, 2017.

\bibitem{Watters:2017}
Nicholas Watters, Andrea Tacchetti, Theophane Weber, Razvan Pascanu, Peter
  Battaglia, and Daniel Zoran.
\newblock Visual interaction networks: Learning a physics simulator from video.
\newblock {\em NIPS}, 2017.

\bibitem{Zheng:2018}
David Zheng, Vinson Luo, Jiajun Wu, and Joshua~B. Tenenbaum.
\newblock Unsupervised learning of latent physical properties using
  perception-prediction networks.
\newblock {\em UAI}, 2018.

\bibitem{Sanchez-Gonzalez:2018}
Alvaro Sanchez-Gonzalez, Nicolas Heess, Jost~Tobias Springenberg, Josh Merel,
  Martin Riedmiller, Raia Hadsell, and Peter Battaglia.
\newblock Graph networks as learnable physics engines for inference and
  control.
\newblock {\em ICML}, 2018.

\bibitem{Hamrick:2017}
Jessica~B. Hamrick, Andrew~J. Ballard, Razvan Pascanu, Oriol Vinyals, Nicolas
  Heess, and Peter~W. Battaglia.
\newblock Metacontrol for adaptive imagination-based optimization.
\newblock {\em ICLR}, 2017.

\bibitem{Pascanu:2017}
Razvan Pascanu, Yujia Li, Oriol Vinyals, Nicolas Heess, Lars Buesing, Sebastien
  Racanière, David Reichert, Théophane Weber, Daan Wierstra, and Peter
  Battaglia.
\newblock Learning model-based planning from scratch.
\newblock {\em arXiv:1707.06170}, 2017.

\end{thebibliography}

\end{document}